\documentclass[letterpaper, 10 pt, conference]{ieeeconf}
\usepackage{amsmath,epsfig,subfigure}

\title{\LARGE \bf Collective Energy Foraging \\of Robot Swarms and Robot Organisms}

\author{Serge Kernbach \\
{\small Institute of Parallel and Distributed Systems, University of Stuttgart} \\
{\small Universit{\"a}tsstr.~38, D-70569 Stuttgart, Germany}, {\small \it Serge.Kernbach@ipvs.uni-stuttgart.de}
}

\begin{document}

\maketitle
\thispagestyle{empty}
\pagestyle{empty}

\begin{abstract}
Cooperation and competition among stand-alone swarm agents increase collective fitness of the whole system. A principally new kind of collective systems is demonstrated by some bacteria and fungi, when they build symbiotic organisms. Symbiotic life forms emerge new functional and self-developmental capabilities, which allow better survival of swarm agents in different environments. In this paper we consider energy foraging scenario for two robotic species, swarm robots and symbiotic robot organism. It is indicated that aggregation of microrobots into a robot organism can provide better functional fitness for the whole group. A prototype of microrobots capable of autonomous aggregation and disaggregation are shown.
\end{abstract}

\section{INTRODUCTION}

Natural collective systems demonstrate that many individuals cooperate when this is profitable to each of them. Examples are cooperative hunting of predatory animals, group-based foraging of mammals~\cite{Stephens87} or nest building of social insects~\cite{Bonabeau99}. In these and many other examples, animals get together when it provides better chances for foraging, for defence, or generally for surviving in their environment. Participants of these groups can be weak with limited sensors/actuator capabilities, however collectively they can build a strong group with very extended capabilities.

Lately, technical systems mimic natural collective systems in improving functionality of artificial swarm agents. Collective, networked or swarm robotics are scientific domains, dealing with a cooperation in robotics~\cite{Sahin04}, \cite{Kernbach11-HCR}. Research in collective robotics is mostly concentrated on stand-alone autonomous robots. Cooperation and competition among stand-alone robots increase their collective fitness~\cite{Kornienko_S06}. However, natural swarm agents can build a principally new kind of collective systems. For example, fungi \emph{dictyostelium discoideum} can aggregate into a multi-cellular symbiotic organism and perform such activities that cannot be fulfilled alone or in a swarm-like way~\cite{Haken78-83}. The symbiotic organization emerges new functional capabilities, which allow swarm agents to achieve better fitness in the environment~\cite{Levi10}. When the need of aggregation is over, symbiotic organism can disaggregate and exists further as stand-alone agents~\cite{Kernbach08_2}.

Swarm robots can also build symbiotic life forms and achieve better functional fitness. To demonstrate this idea, we consider a collective energy foraging scenario for microrobots Jasmine~\cite{Kornienko_S05d}. Swarm robots can autonomously find an energy source and recharge. Dedicated collective strategies~\cite{Kernbach09Nep} can essentially improve the efficiency of energy foraging, but nevertheless a functional fitness of a swarm is limited. For instance, if the recharging station is separated from a working area by a small barrier, robots can never reach the energy source. However, when robots will aggregate into more complex symbiotic organism, which can pass the barrier, they will reach the docking station. In this paper we demonstrate main differences in hardware and software as well as in behavioral strategies of both robot systems: swarm robots and symbiotic organisms. Based on the existing microrobotic platform Jasmine IIIp~\cite{Swarmrobot2011}, a prototype of multi-robotic organism is developed. A few topological and functional issues of aggregation into an organism are shown~\cite{Kernbach08Permis}.

One of the main arguments for the collective energy foraging is related to the idea of functional self-development: robot organisms can autonomously change their functionally. The energy homeostasis can appear as a task-unspecific criterion, which defines such self-development. Since energetic efficiency, by analogy to natural survival, possess open-ended features, the collective energy homeostasis can be considered as a candidate underlying a design of open-ended self-developmental systems~\cite{Kernbach08online}.

The rest of the paper is organized in the following way. In Sec.~\ref{sec:foraging} the energy foraging scenario and the corresponding hardware and software of swarm robots are described. Sec.~\ref{sec:strategy} demonstrates collective strategies and limitations imposed on collective fitness of the robots. Sec.~\ref{sec:strategy} briefly overviews the development of symbiotic microrobots. Finally, in Sec.~\ref{sec:conclusion} we conclude this work.

\section{ENERGY FORAGING IN A ROBOT SWARM}
\label{sec:foraging}

The distinctive property of any living organisms is the energy homeostasis and, closely connected, foraging behavior and strategies~\cite{Stephens87}. The robots, equipped with on-board recharging electronics, can also possess its own energy homeostasis~\cite{Silverman02}. In this way, when swarm robots become "hungry", they can collectively look for energy resources and execute different strategies in cooperative energy foraging~\cite{Kornienko_S06b}. In critical cases robots can even decide to perform individual foraging, competing with other robots for resources. In this case we also see analogies to behavioral strategies of animals~\cite{Camazine03}.

The need of energy is a perfect example of natural fitness. When robots, with their strategies, found enough energy, they survive in the environment. It means these strategies are good enough, other case robots energetically die. Based on energy foraging, several evolutionary approaches for different robot species can be developed, compared and tested. Further in the section we briefly demonstrate the developed solutions for swarm energy foraging and show a limit of functional fitness in a swarm.

\subsection{RECHARGING HARDWARE}
\label{sec:hardware}

The hardware is described in ~\cite{Kornienko_S06b}, \cite{Kernbach09Nep} and generally follows the idea of swarm embodiment~\cite{Kornienko_S05e}; here we briefly overview it. To make the microrobots capable of autonomous recharging, four following components are required: (1) internal energy sensors, monitoring energy level of Li-Po accumulator; (2) especial recharging circuits for Li-Po process; (3) reliable connectors to docking stations with a low electrical resistance; (4) communication with docking stations. Internal energy sensor is implemented as a resistive voltage divider with the coefficient 0.55, it consumes a continuous current of 5$\mu$A. The voltage divider is connected to a non-regulated power line and with an ADC of the microcontroller.

\begin{figure}[ht]
\centering
\subfigure[]{\includegraphics[width=.21\textwidth]{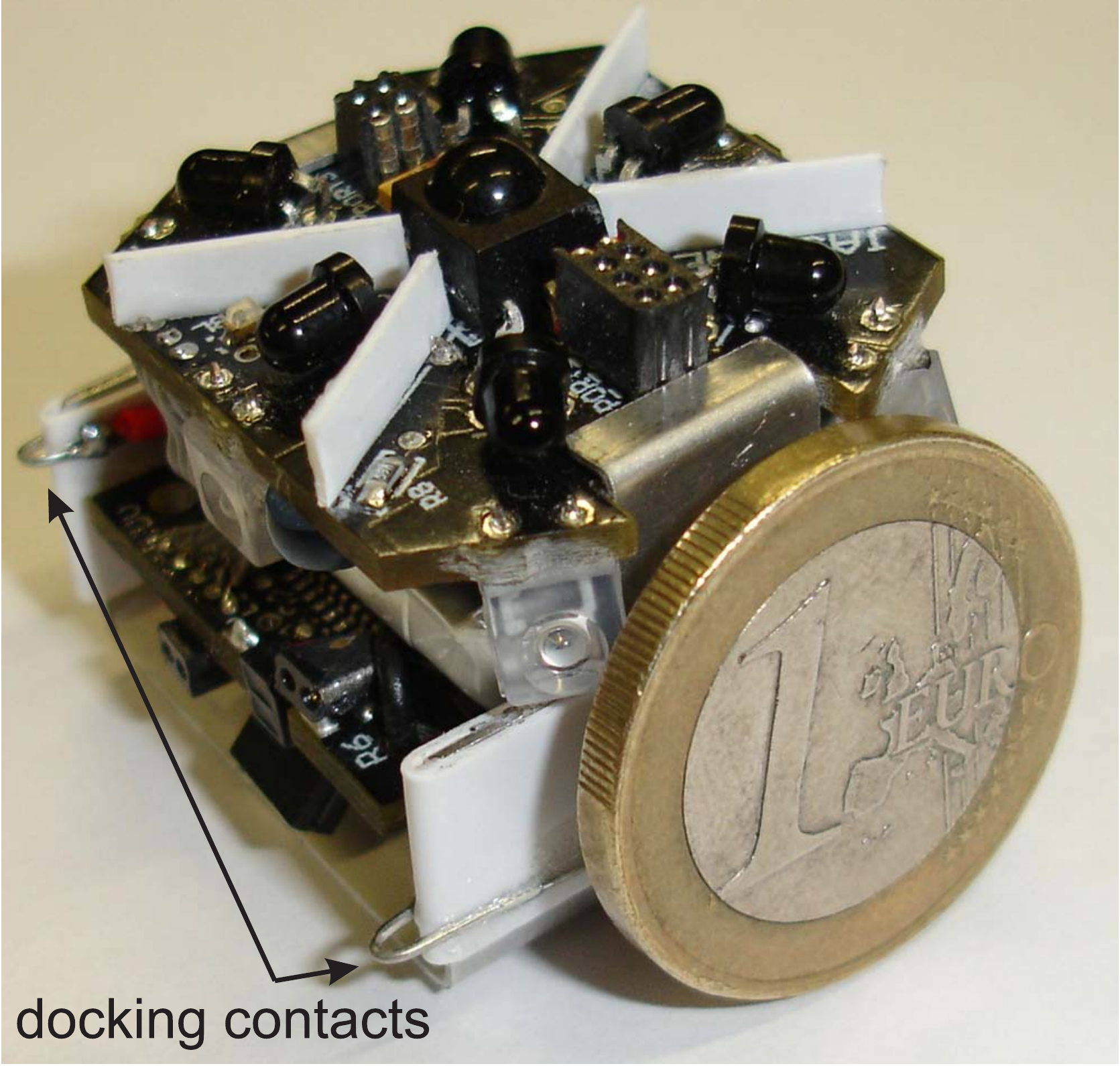}}~
\subfigure[]{\includegraphics[width=.27\textwidth]{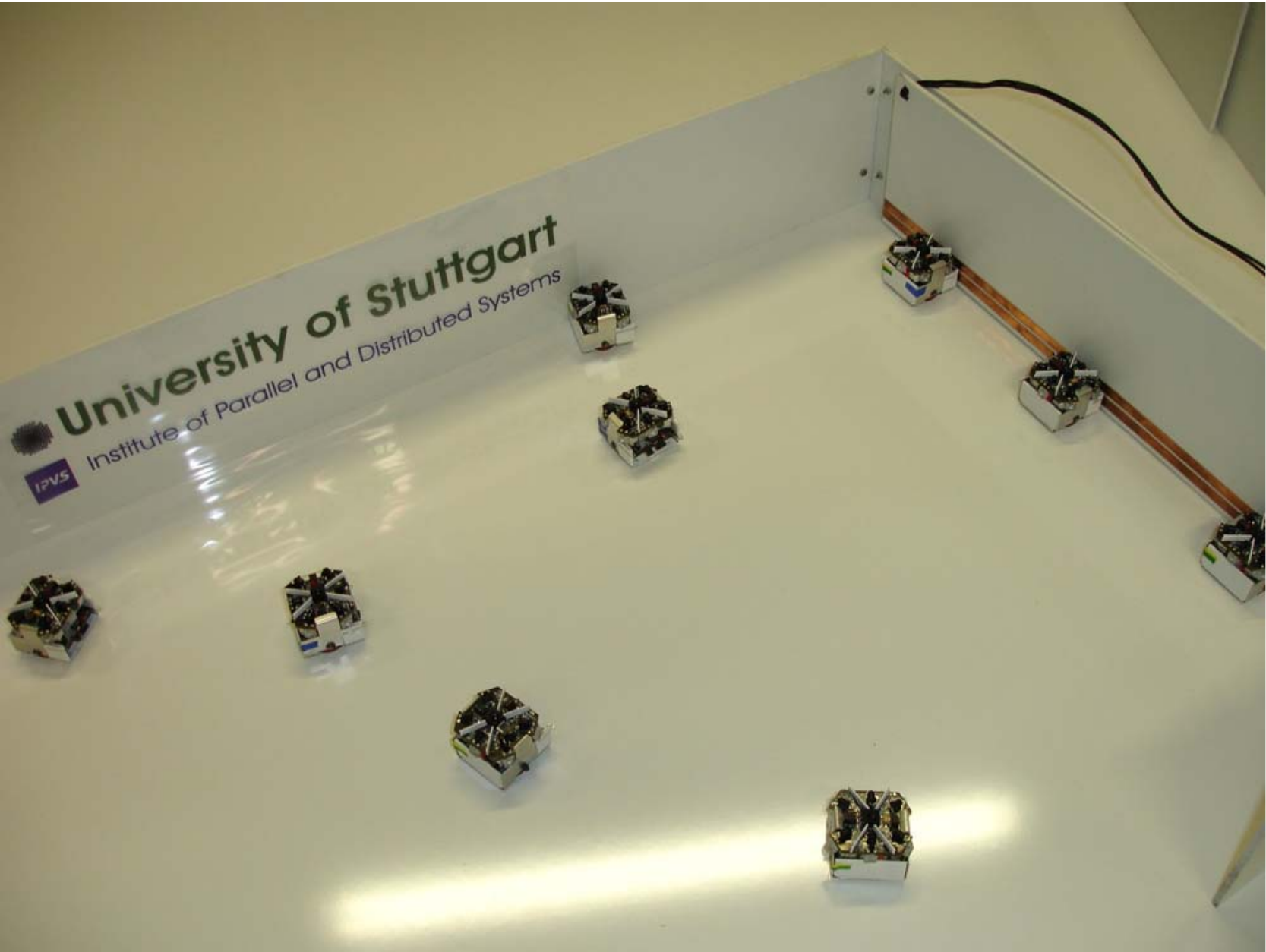}}
\caption{\small \textbf{(a)} Microrobot Jasmine IIIp, shown are recharging contacts; \textbf{(b)} The docking station and a few robots recharging there. Images are from \cite{Kornienko_S06b}. \label{fig:robot}}
\end{figure}

The fully charged Lithium-Polymer battery provides the micro-robot with energy for nearly 1,25 hours. Li-Po accumulators require specific recharging process. To control it we use the LTC4054 Li-Ion/Li-Po battery charger. It is small, with a low dropping voltage and can be directly installed on the Li-Po accumulator. To perform autonomous recharging, robots have to possess connectors to the docking station, which allows simple docking and reliable mechanical contact of low electrical resistance. After several tests~\cite{Jebens06} the best solution was two 0.4mm silver-plated wires glued in the front area of the robot. Both connectors are installed on different height, see Fig.\ref{fig:robot}(a).
The docking station represents a wall with glued 0.2mm thin copper stripes of 5mm wide, see Fig.~\ref{fig:robot}(b). Both copper stipes are connected to the 5V source. Many such docking stations can be placed together, see
Fig.~\ref{fig:docking}. To connect to the docking station, a robot has to move to this wall (based on the docking signal), until it gets a positive signal from the touch sensor. After that the robot shortly turns on its wheels to produce a small mechanical strain. The communication system of the docking station sends continuously the signal "I'm free slot in the docking station" (coded numerically). This signal can be received within 10-15cm away from the docking station. Discharged robots, when sensing a free docking slot, approach the docking station and start a recharging process. Body of the recharging robot blocks the signal, so that no other robot can receive the signal from this slot. When recharging is finished, the robot moves away.

\subsection{ENERGETIC HOMEOSTASIS OF SWARM ROBOTS}
\label{sec:homeostasis}

As described in the previous section, the robot is able to sense its own energetic level and the position and availability of energy sources. Capacity of a single cell Li-Po accumulator is enough for a running time of at least 1,25 hour. The optimal working mode of Li-Po accumulator is discharging only up to 75-80\% of capacity. The critical level of accumulator is the voltage dropping less than 3V, because in this case the internal power regulator is not able to stabilize voltage fluctuations and microcontroller can spontaneously reboot. The full recharging takes about 90 minutes, the partial recharging is almost equal to discharging (15 min. motion requires about 15 min. recharging).

The energetic homeostasis of the robot is related to the voltage level of the Li-Po accumulator and includes five different states. When the voltage drops under 3.05V, a robot should stop and go in stand-by mode. In this state it is not able to react on external stimulus and needs a human assistance for recharging. When the voltage is under 3.2V, a robot should look for docking station independently of the current task. It has about 3-5 minutes to find it, other case it will energetically die. When the voltage is less than 3.7V but more than 3.2V, the robot has different degrees of discharging. It means a robot can start look for energy sources, when there are no more important tasks. The lower the energetic level is, the higher is the priority of looking for energy. Ideally, when reaching 3.65V, robot should start looking for the docking station. When during recharging a robot achieves 4.0V, this state can be characterized as satisfied -- accumulator is not fully recharged (80-85\%), but enough to run again and make slot free for other robots. The voltage increases between 4.1 and 4.2 very slow, (accumulator is already recharged up to 90-95\%). When voltage is about 4.2V (ADC value 196) the circuit stops recharging. In this state the robot is fully recharged.

Firstly, in a critical state, robots should break the currently executed collective or individual activity. This is not typical in robotics, however a died robot can essentially distort collectively executed activity. Secondly, a robot should have the priority of a currently executed activity and the priority of looking for energy sources. When, for example, the priority of current activity is 0.6, but the level of discharging is 0.7, robot will look for a docking station. Finally, a robot can have so-called "collective instinct", it can recharge only till "satisfied state" (it takes less time), and makes a slot free for recharging of another robot.

Collective energy foraging can be designed on the one hand as an artificial self-organizing process~\cite{Kernbach08}, which possess adaptive~\cite{kernbach09adaptive} or controllable-emergent~\cite{Kornienko_S04b} properties. On the other hand, application of planning-based approaches, e.g.~\cite{Kornienko_S03A}, \cite{Kornienko_S04}, is also possible. Generally it depends on cognitive capabilities of the robot~\cite{Kornienko_S05a}, in particular on the sensing system~\cite{KornienkoS05d}. Independently of the selected approach, a cooperativeness of individual energetic homeostasis consists in management of (a) a priority of collective tasks, (b) robots with critical energy states or (c) the regarding behavior.

\section{COLLECTIVE ENERGY FORAGING AND LIMITATION OF FUNCTIONAL FITNESS}
\label{sec:strategy}

The need of collective strategy in energy foraging appears from the optimization of the swarm efficiency. The energetic swarm efficiency $\Phi$ is calculated as a relation of the time spend for useful activities $t_{TASK}$ and the total time $t_{TASK} + t_{RECHARGING}$. Remembering that the working time is equal to the recharging time, the best swarm efficiency is
\begin{equation}
\label{eq:efficiency}
\Phi=\frac{t_{TASK}}{t_{RECHARGING}+t_{TASK}}=0.5.
\end{equation}
This best efficiency is approximatively achievable when one robot works alone ($\Phi=0.45-0.48$). When there are many robots, a few undesired effects can appear: (1) the docking station can be a "bottle neck" that essentially decreases the swarm efficiency; (2) robots with a high-energy level can occupy the docking station and block low-energetic robots. These robots can energetically die (and so decrease the swarm efficiency); (3) many robots can create a "crowd" around a docking station and essentially hinder a docking approach. This can increase the total recharging time and makes worse the energetic balance of the whole swarm.

Robots, in their energetic homeostasis, have only two possible individual decisions: to execute a current collective task or to move for recharging. In this way, a cooperative strategy should find a right timing and a right combination between these individual decisions of all robots.
\begin{figure}[htp]
\centering
\subfigure[]{\includegraphics[width=.49\textwidth]{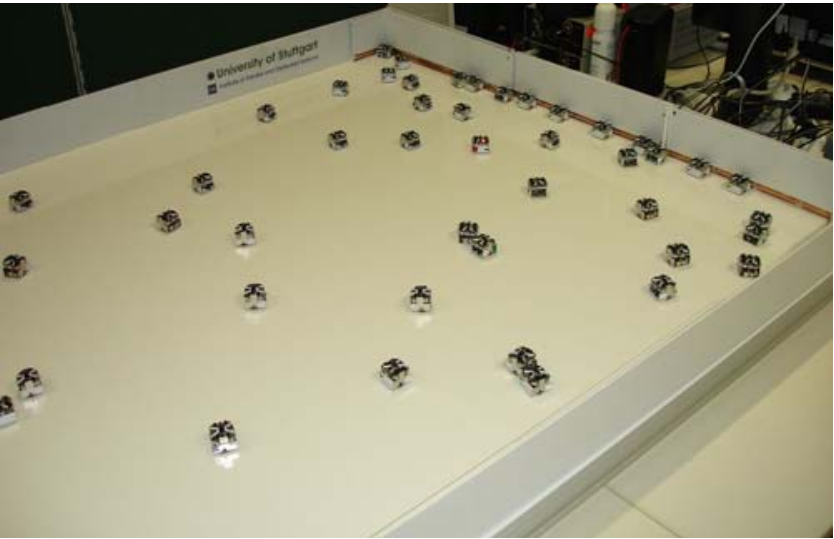}}
\subfigure[]{\includegraphics[width=.49\textwidth]{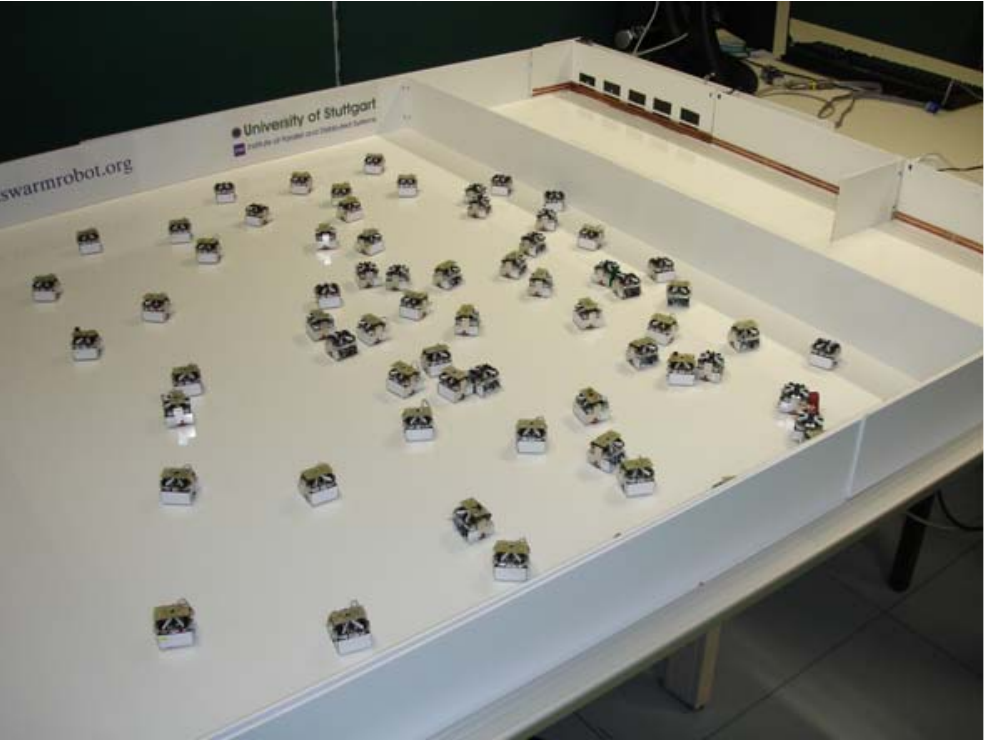}}
\subfigure[]{\includegraphics[width=.49\textwidth]{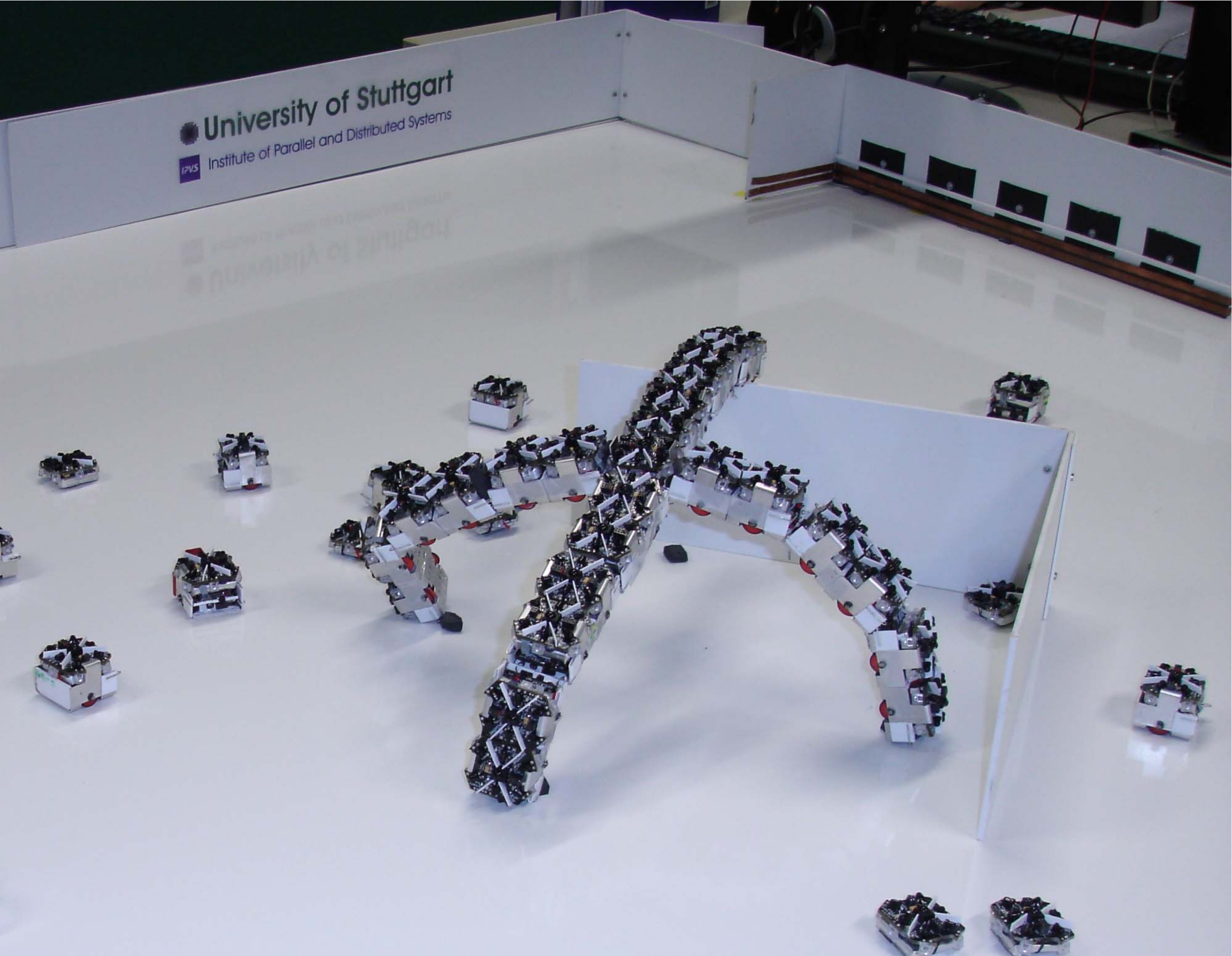}}
\caption{\small \textbf{(a)} Docking of a few robots for recharging. Shown is the two-line approach: the first line - recharging robots, the second line - robots waiting for recharging; \textbf{(b)} The "barrier problem" - robots are separated form docking stations by a barrier; \textbf{(c)} A possible solution to the "barrier problem": swarm robots form a multi-robot organism and collectively pass the barrier.\label{fig:docking}\label{fig:wallPassing}}
\end{figure}

There are two ways to synchronize individual decisions. Firstly, the decision making procedure can be done on the collective level and all robots have only to execute the obtained decision~\cite{Kornienko_OS01}. Examples of such a decision making process are bargaining, auctioning or similar, where the final collective activity is not fixed and should first be negotiated~\cite{Weiss99}. Application of numerical approaches for collective decision making is also possible~\cite{Levi99}. Secondly, the decision can be done individually, however input information for this decision will be prepared collectively. This decision making process is usually applied for "switching decision", where the final collective activity represents a sequence of predefined sub-activities. Obviously, the individual state of a robot also influences its own decision, so that finally a collective behavior represents a complex combination between collective "needs" and individual "desires". Lately, several strategies of energy foraging for a robot swarm up to 50 swarm agents are implemented, see Fig.~\ref{fig:docking}. These cover different bio-inspired approaches~\cite{Habe07},~\cite{Kancheva07} and hand-coded strategies~\cite{Attarzadeh06}. The efficiency of these approaches is between 0.18 (mostly hand-coded) and 0.33 (bio-inspired).

In one of these experiments, a few robots died nearly the docking station, so that they build a barrier from one border to other one in the robot arena (we "simulated" this in the Fig.~\ref{fig:docking}(b)). Robots located before this barrier (away from the docking station) will energetically die. This is the limit of a functional fitness of swarm robots. There is no strategy that allow swarm robots to overpass the barrier. Only when swarm robots will collectively emerge new functionality, such as "pass the barrier", they will solve the "barrier problem". In the next section we consider this case.

\section{ROBOT ORGANISMS}
\label{sec:symbiotic}

An approach to solve the "barrier problem" is an aggregation into a multi-robot organism. They can achieve the docking stations by "growing legs" and overstepping the barrier. So the robots are helping each other in a symbiotic life form, see Fig.~\ref{fig:wallPassing}(c). Obviously, such a solution is extremely challenging from many viewpoints. Symbiotic robot systems have many similarities with known robotic research as e.g. mechanical self-assembling~\cite{Ishiguro06} or reconfigurable robotics~\cite{Murata06}: (1) Robots should be capable for autonomous aggregation and disaggregation; (2) Robots in the disaggregate state should possess individual locomotion; There is no central control neither for disaggregated state (swarm) nor for the aggregate state (organism), it should possess a kind of guided self-organization~\cite{Kornienko_S04a}; (4) Stand-alone robots should profit from the aggregation into organism.

\subsection{BRIEF OVERVIEW OF HARDWARE AND SOFTWARE}
\label{sec:symbioticHardware}

\begin{figure}[htp]
\centering
\subfigure[]{\includegraphics[width=.25\textwidth]{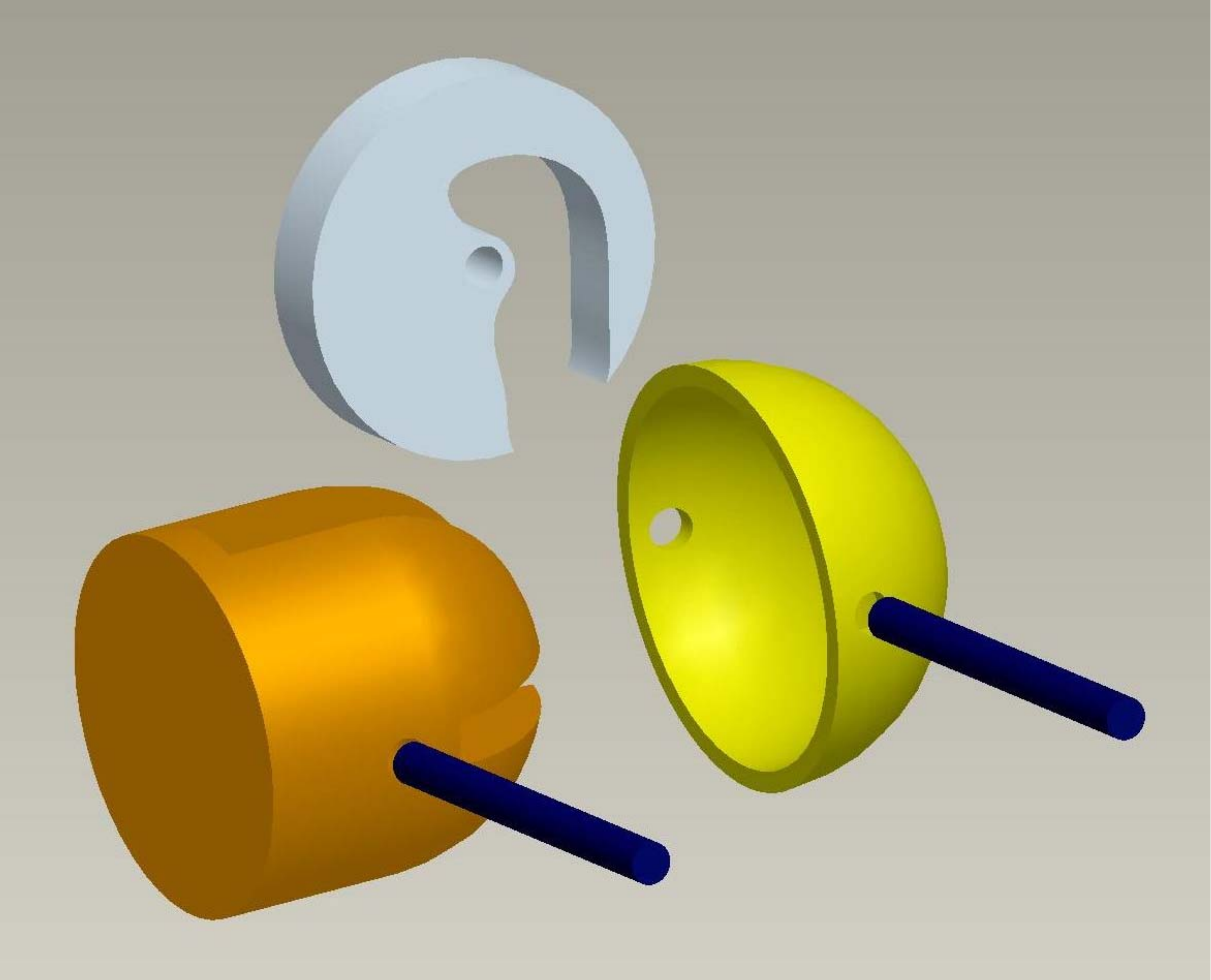}}~
\subfigure[]{\includegraphics[width=.22\textwidth]{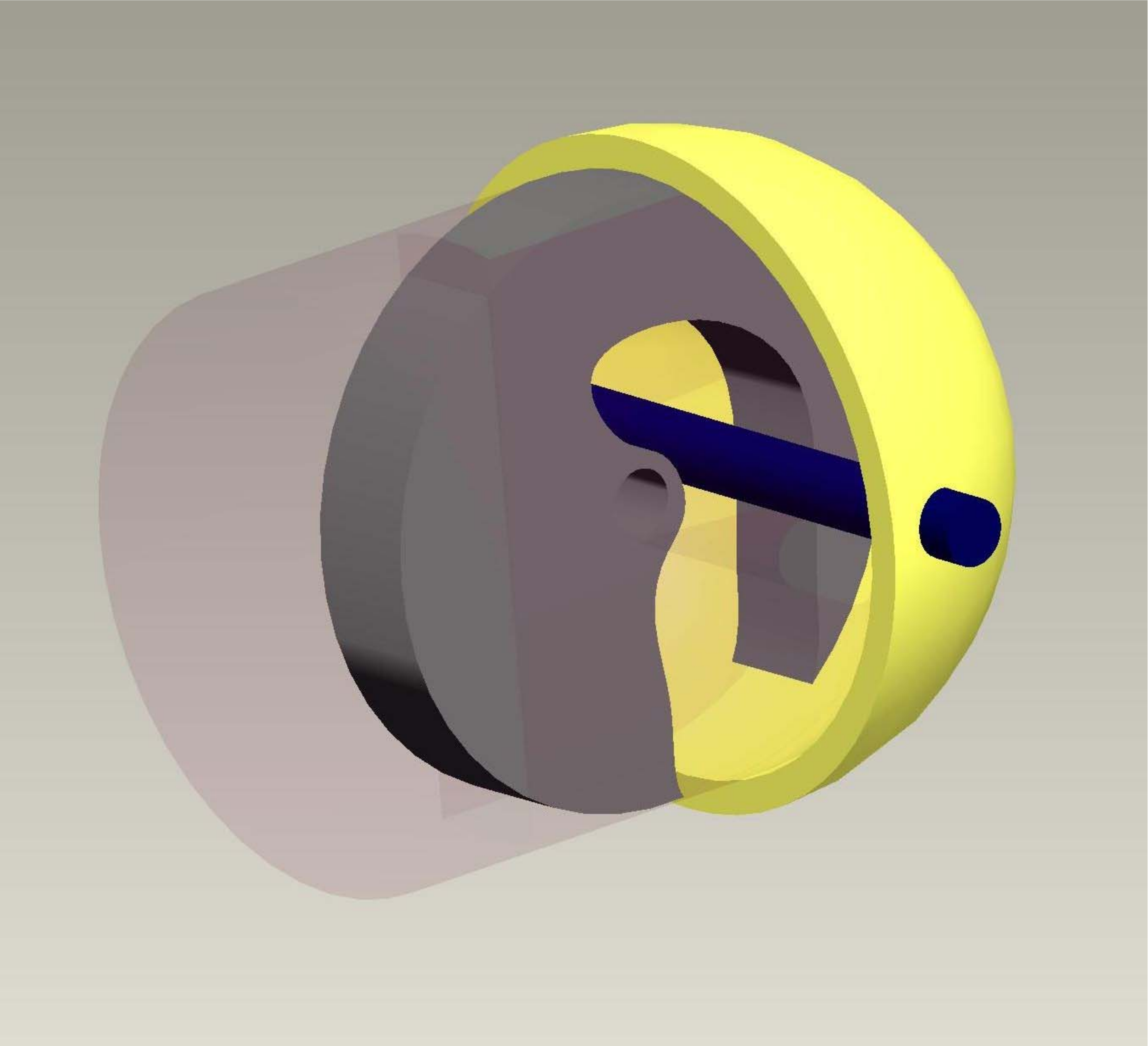}}
\subfigure[]{\includegraphics[width=.49\textwidth]{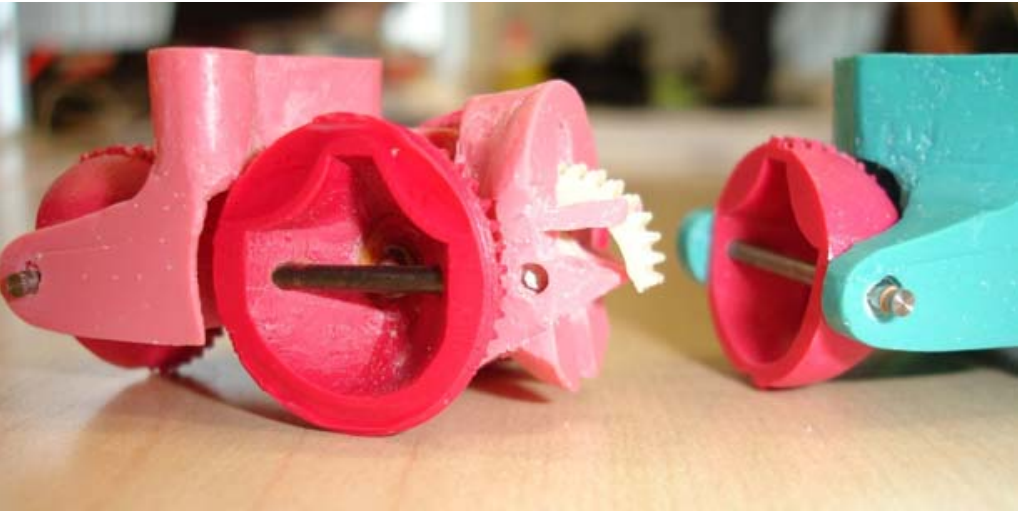}}\\
\subfigure[]{\includegraphics[width=.24\textwidth]{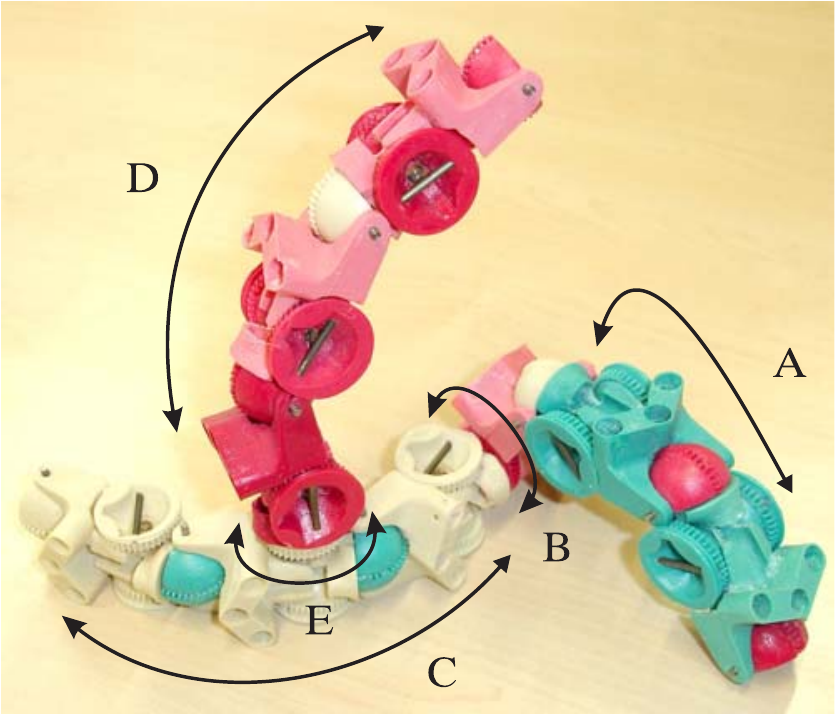}}~
\subfigure[]{\includegraphics[width=.24\textwidth]{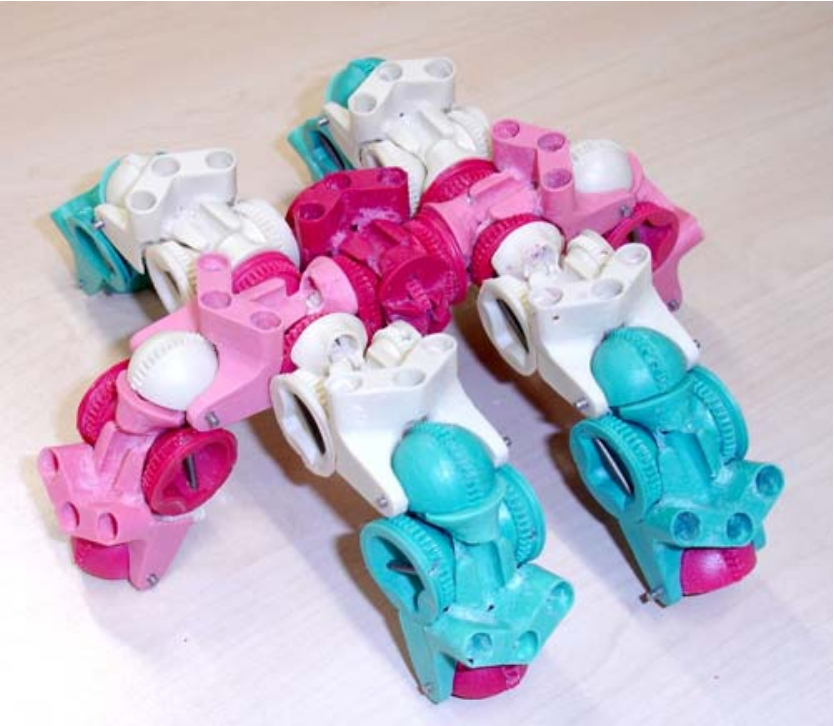}}
\subfigure[]{\includegraphics[width=.48\textwidth]{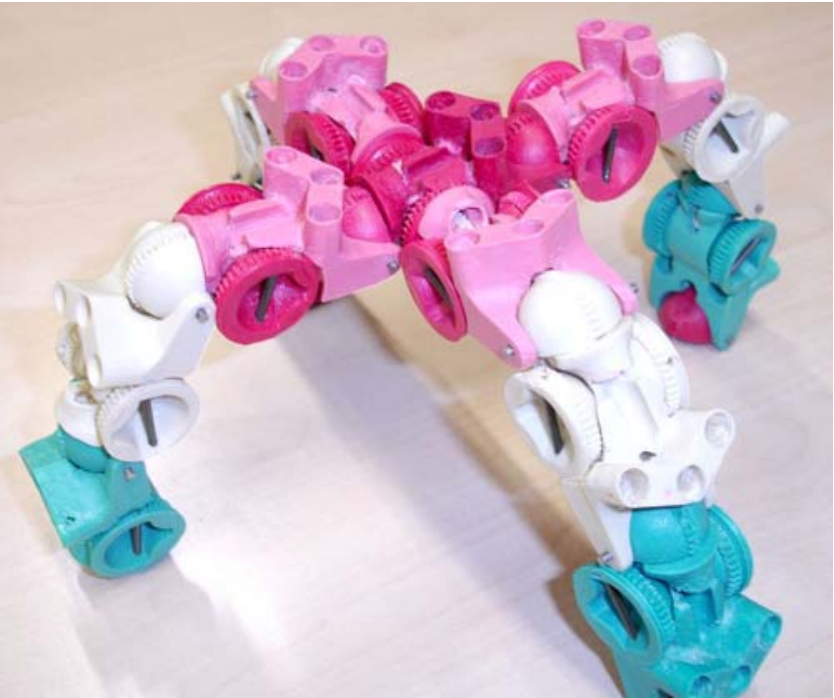}}\\
\caption{\small \textbf{(a)} 3D Assembling diagram of the docking mechanism; \textbf{(b)} Lock with a hook; \textbf{(c)} Implementation of the docking mechanism in the prototype; \textbf{(d)} 5 degrees of freedom for the organism when using two rotational connectors; \textbf{(e),(f)} Different locomotion principles, obtained by combining several DOF; \label{fig:dockMechanism} \label{fig:DOF}\label{fig:LocomotionOrganims} }
\end{figure}

This section demonstrates a developed solution before start of the projects~\cite{symbrion}, \cite{replicator} within so-called internal pilot projects. This version of robots is based on the Jasmine III+ development, shown in Fig.~\ref{fig:robot}(a). From the hardware viewpoint, the most challenging part is to provide simple and reliable docking mechanism, which allows an autonomous aggregation and disaggregation. The successful solution is shown in Fig.~\ref{fig:dockMechanism}.
The docking mechanism consists of one male part, placed on the front of the robots and three female parts, placed in wheels and on the back, see Fig.~\ref{fig:dockMechanism}(c). Male and female connectors can be locked with a hook, as shown in Fig.~\ref{fig:dockMechanism}(b). This mechanism is experimentally tested in a small group of 13 prototype robots. Combining different wheel- and back- connectors, various configurations with many degrees of freedom (DOF) can be obtained, see Fig.~\ref{fig:DOF}(d). Combining several DOF, the organism can demonstrate different locomotion principles, as shown in Fig.~\ref{fig:LocomotionOrganims}(e),(f).

\begin{figure}[ht]
\begin{center}
\subfigure[]{\includegraphics[width=.48\textwidth]{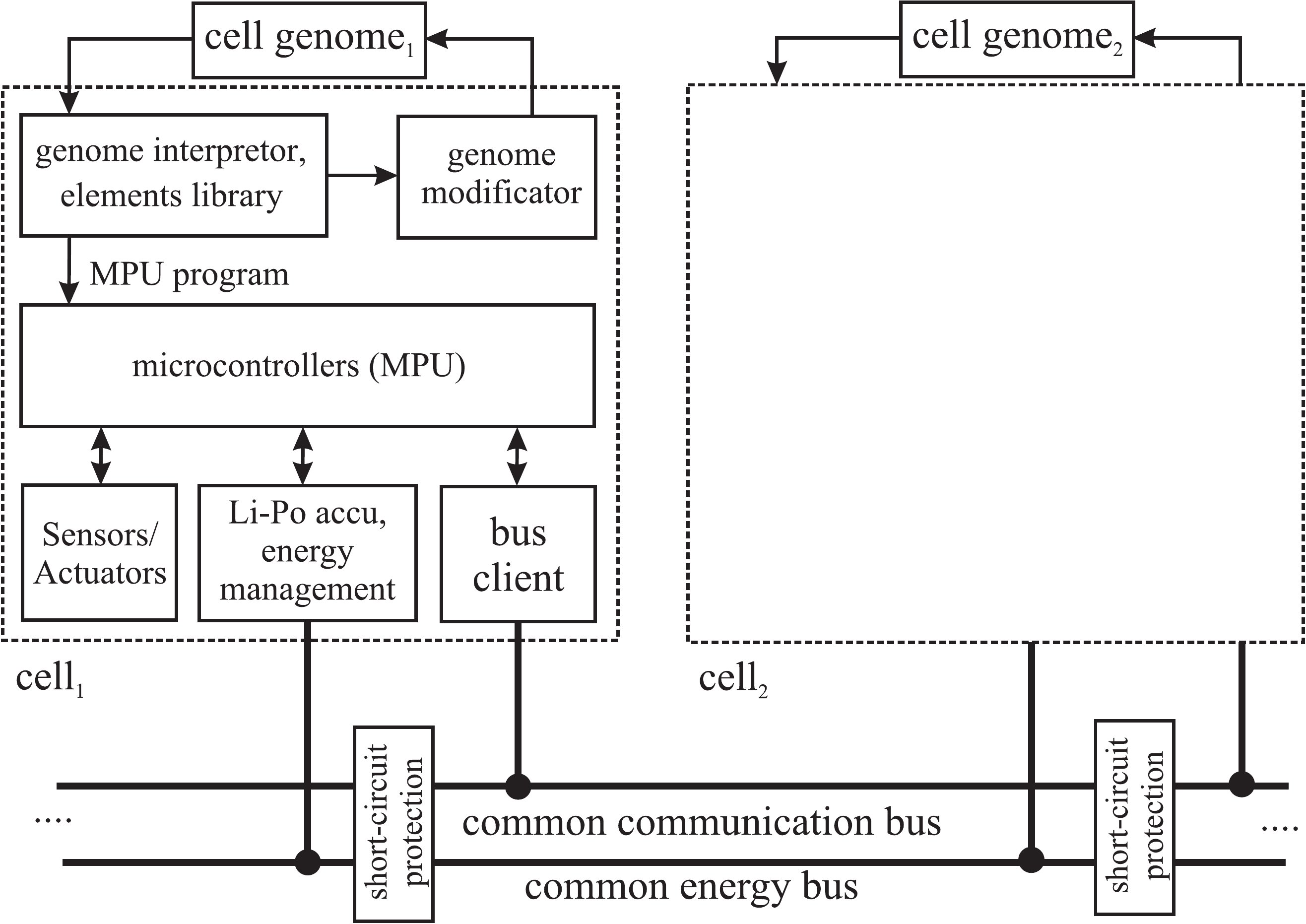}}
\subfigure[]{\includegraphics[width=.45\textwidth]{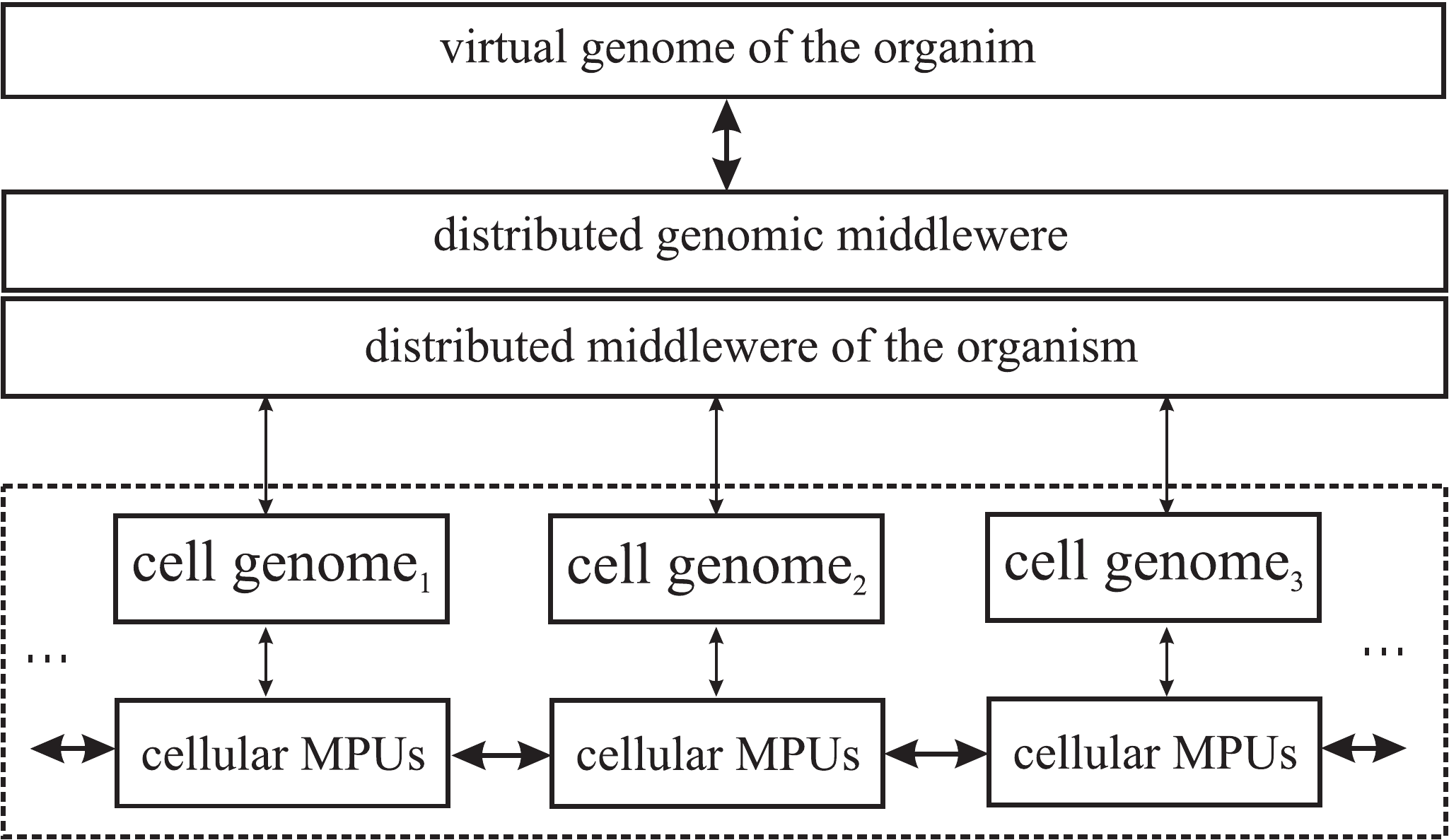}}
\caption{\small  \textbf{(a)} The functional structure of the symbiotic organism; \textbf{(b)} Computational structure of the symbiotic robot organism. \label{fig:functional} \label{fig:framework}}
\end{center}
\end{figure}

The functional structure of the organism is shown in Fig.~\ref{fig:functional}(a). The organism should possess a common energy and communication bus. Client electronics for both busses should allow taking and giving energy from/into the bus and a high speed serial data transfer. Special short-circuit protection switches off such segments of energy and data busses that are defect. In this way, even when some robots cause failure on central busses, the organism will survive. Each robot has 250mA/h Li-Po accumulator connected in parallel in the organisms (with Li-Po balancer). When organism has about 30 robots, common energy resources are about 7,5A/h. With even 4C short-time discharging, the organism can produce about 30A current for application purposes (intended size of the organism is about $30\times30\times30$cm and wight about 500gr.). When taking into account that only a few robots in the organism perform actuation (and so have a maximal energy consumption), the aggregation into the organism should be profitable for robots at least from the energetic viewpoint.

In the current development, each robot is controlled through three Atmel microcontrollers (MCU), which communicate via internal I$^2$C bus: MPU with external memory for behavioral programs, MPU for control of sensors/actuators and MPU for ZigBee and IR communication. Computational resources of the organism are separated on three layers: physical, middleware and virtual genomic one, see Fig.~\ref{fig:framework}(b). MPUs communicate via data bus and build a distributed computational system. Each robot supports middleware, which allows running user-defined tasks on many MPUs in parallel and uses distributed memory resources. The middleware system takes technical care about introducing new members into the organism or excluding some robots from the organism.

The main MPU is programmed by so-called genetic framework~\cite{Nagarathinam07}, \cite{Kernbach08online}, \cite{Schwarzer08}. A genome of an artificial organism carries the total set of genes and includes instructions for building, running and maintaining the organism. Genes are functionally complete and consist of small 'states' (by analogy to biological 'bases'~\cite{Alberts02}). For example, the function 'move' is a gene and consists of different states that control a movement. Using these states, it is possible for genes as well as for the whole genomes to be manipulated through recombination or mutation. Genome is stored in external nonvolatile EEPROM memory and has a maximal size 2$^{16}$ states. Thus, when a robot is switched off, it does not lose its own information. The function of gene modification is accelerated by the hardware. In this way, genomes of individual robots, through middleware system, build virtual genome of the organism.

\subsection{ENERGY HOMEOSTASIS OF A ROBOT ORGANISM}
\label{sec:symbioticHomeostasis}

The preliminary experiments with symbiotic organisms allowed us to detect a few critical issues for further development. First of all, when robots aggregate into an organism, they save energy. From this point of view, symbiotic life form is more profitable for robots. The targeted energy homeostasis has the form similar to one, discussed in Sec.~\ref{sec:homeostasis}, only on the level of the whole organism. It is intended the preprogrammed middleware takes care about internal energy management, whereas the genome controls behavioral strategies of finding energy sources. The organism should develop these strategies evolutionary.

Swarm robots, when moving on the arena, have preprogrammed basic genes, such as "move", "rotate", "dock in", "dock from", "actuate" and others. All basic genes and their combinations have two weights: "success" and "potentially consumed energy". All these genes build a search space in EEPROM. The gene sequences which leaded to successful recharging are more weighted in "success". When some combination does not lead to a success, its "success" importance is rapidly decreased. After each successful recharging, a robot updates "consumed energy". It is also possible that robots have a logbook of all updates for gene sequences.

\begin{figure}[ht]
\centering
\includegraphics[width=.45\textwidth]{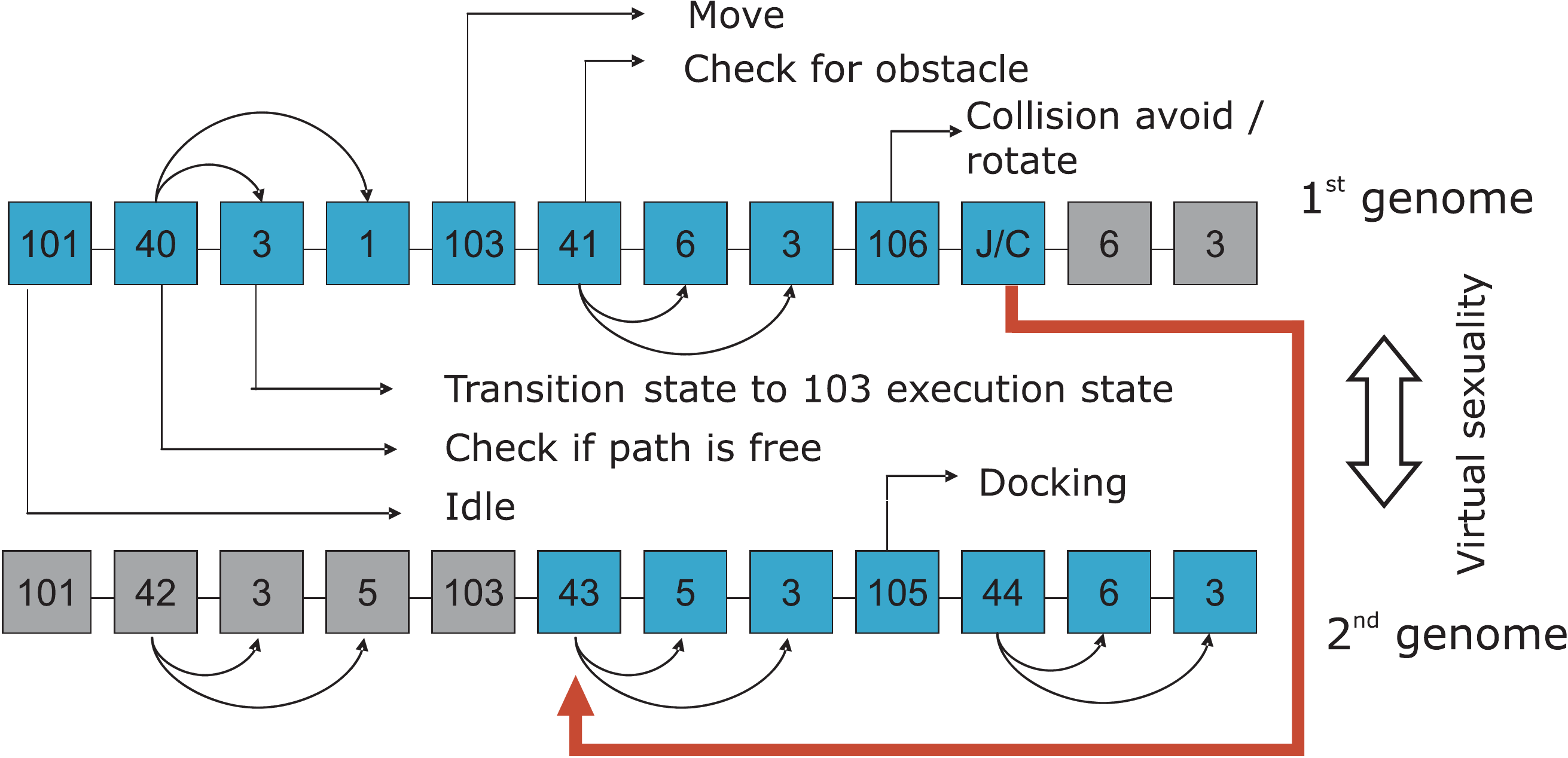}
\caption{\small Genome Mutation-Adding/overwriting via virtual robot sexuality (from~\cite{Nagarathinam07}).
\label{fig:mutation2}}
\end{figure}

Swarm robots firstly combine low-energy sequences, such as "move" or "rotate". When these low-energy sequences have zero-near success weight (e.g. the barrier problem"), robots look for further combinations of basic commands, which have a higher energy consumption. In this way several robots can aggregate together. In this aggregation step we encountered the first problem. All robots have different genomes. Their simple combinations did not lead to useful solutions~\cite{Nagarathinam07}, \cite{Koenig07_2}, \cite{Speidel08}. Robots in the organism start to develop their common virtual genome and only then to introduce into this common genome their individual information as a sequence of recessive genes. The problem is even harder, when a robot already has a genome from some past symbiotic forms. Currently it is not clear whether all symbiotic robots converge to an unique genome with different recessive sequences. One of the concept in this way is to use the "virtual robot sexuality" -- robots, even during disaggregated phase, exchange genetic information~\cite{Schwarzer08}, see Fig.~\ref{fig:mutation2}.

Another encountered problem is a delayed fitness response of the robot organism. Typical life-cycle of swarm robots on the 110$\times$140cm arena is about a few minutes. The negative feedback is issued, when during this time no energy source is found. The life-cycle of the symbiotic form is much larger. This means that a positive feedback can be achieved only after a large number of unsuccessful steps, some of them can even destroy the organism.

\section{CONCLUSION}
\label{sec:conclusion}

In this short paper we have shown a new paradigm in collective systems, where the swarm robots get capable of self-assembling into a single multi-robot organism. We introduced an energy foraging scenario for both robot species and demonstrated that a transition between collective and symbiotic robot forms represents a hard problem. It involves not only hardware and software
issues, but also very basic questions being also open in biological sense. We developed a prototype of symbiotic robots capable for working in both, swarm and organism, scenarios. Based on these works, fully functional symbiotic robots will be developed~\cite{Levi10}.

\small
\IEEEtriggeratref{36}

\end{document}